\documentclass{article}
\usepackage{spconf,amsmath,graphicx}
\usepackage{array}
\usepackage{bbding}
\usepackage{amssymb}
\usepackage{mathrsfs}        
\usepackage{graphicx}        
\usepackage{amssymb}         
\usepackage{amsmath}         
\usepackage{enumerate}       
\usepackage{booktabs}
\usepackage[colorlinks,
            linkcolor=red,
            anchorcolor=blue,
            citecolor=blue
            ]{hyperref}

\usepackage[font=small,labelsep=period]{caption}
\usepackage{amsthm}          
\usepackage{algorithm}
\usepackage{algorithmic}
\usepackage{subfigure}
\usepackage{mathrsfs}
\usepackage{graphicx}
\usepackage{amssymb} 
\usepackage{amsmath} 
\usepackage{amsthm}
\usepackage{enumerate}
\usepackage[numbers,sort&compress]{natbib}

\usepackage{amssymb}

\title{MVT: Mask Vision Transformer for Facial Expression Recognition \\
in the wild}
\name{Hanting Li, Mingzhe Sui, Feng Zhao\textsuperscript{$\ast$}\thanks{$^\ast$The corresponding author is Feng Zhao}, Zhengjun Zha, and Feng Wu}
\address{University of Science and Technology of China, Hefei 230027, China \\ \{ab828658, sa20 \}@mail.ustc.edu.cn, \{fzhao956, zhazj, fengwu\}@ustc.edu.cn}
\begin{document}
%
\maketitle
\begin{abstract}
Facial Expression Recognition (FER) in the wild is an extremely challenging task in computer vision due to variant backgrounds, low-quality facial images, and the subjectiveness of annotators. These uncertainties make it difficult for neural networks to learn robust features on limited-scale datasets. Moreover, the networks can be easily distributed by the above factors and perform incorrect decisions. Recently, vision transformer (ViT) and data-efficient image transformers (DeiT) present their significant performance in traditional classification tasks. The self-attention mechanism makes transformers obtain a global receptive field in the first layer which dramatically enhances the feature extraction capability. In this work, we first propose a novel pure transformer-based mask vision transformer (MVT) for FER in the wild, which consists of two modules: a transformer-based mask generation network (MGN) to generate a mask that can filter out complex backgrounds and occlusion of face images, and a dynamic relabeling module to rectify incorrect labels in FER datasets in the wild. Extensive experimental results demonstrate that our MVT outperforms state-of-the-art methods on RAF-DB with 88.62\%, FERPlus with 89.22\%, and AffectNet-7 with 64.57\%, respectively, and achieves a comparable result on AffectNet-8 with 61.40\%.

\end{abstract}
\section{Introduction}
\label{Section 1}
\begin{figure}[htb]

  \centering
  \centerline{\includegraphics[width=8.5cm]{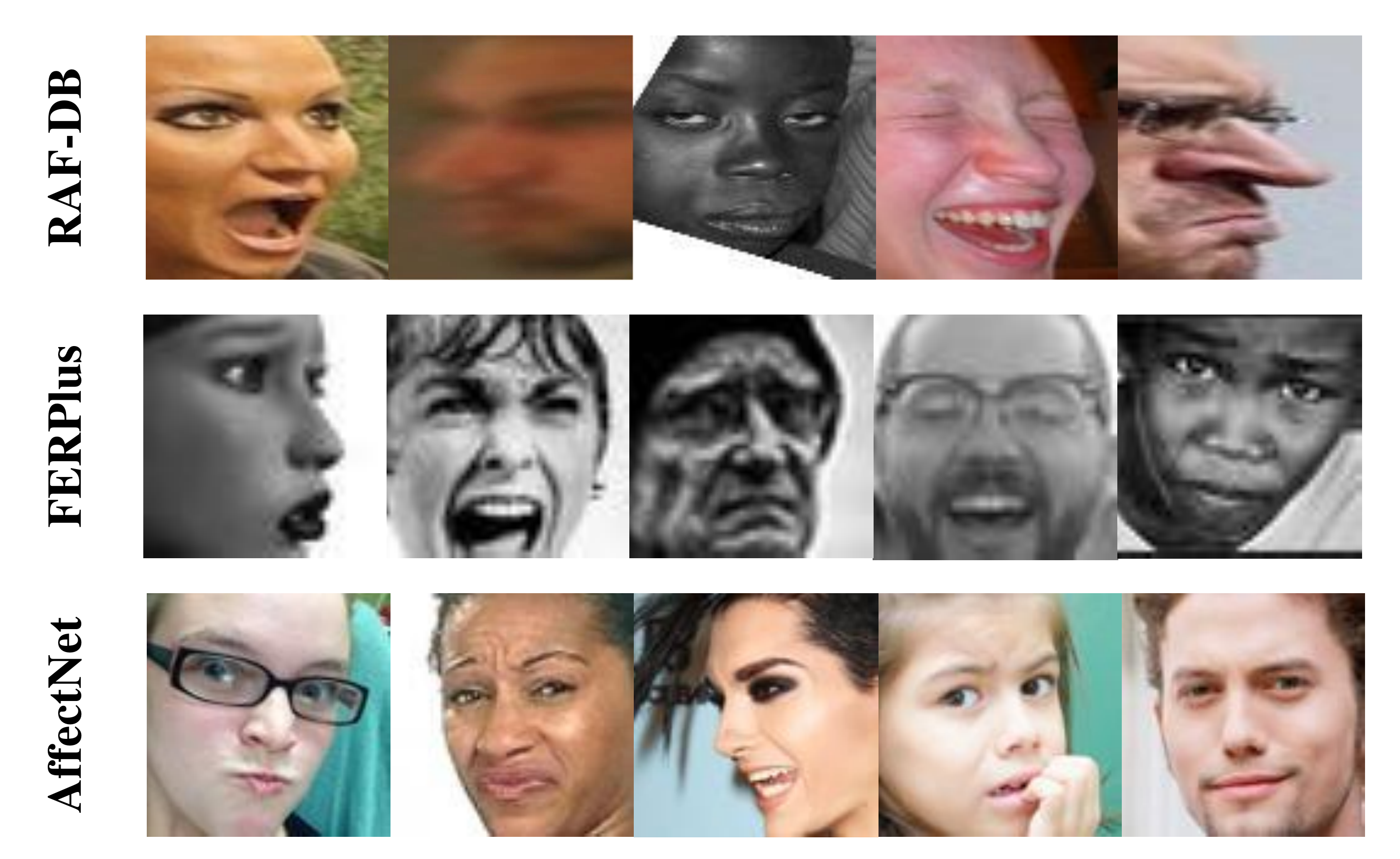}}
  \medskip

\caption{Samples from RAF-DB, FERPlus, and AffectNet. Variant head poses, occlusion,  image quality and backgrounds can bring troubles to FER.}

\end{figure}

Facial expressions are the most natural and direct way for humans to express emotions \cite{tian2001}. Understanding human emotional state is a fundamental premise for many computer vision tasks including human-robot interaction (HRI), driver fatigue monitoring, health-care, etc \cite{zhang2018,li2020survey}. Therefore, many researchers are working on building more robust and stable models to recognize facial expressions. Computer vision tasks usually need large-scale datasets, however, building large-scale laboratorial facial expression datasets requires great costs. Thus, researchers have made significant progress on building FER datasets in the wild, such as RAF-DB \cite{li2017rafdb}, FERPlus \cite{barsoum2016ferplus}, AffectNet \cite{mollahosseini2017affectnet}, EmotionNet \cite{fabian2016emotionet}, etc.

Nonetheless, as illustrated in Fig. 1, FER datasets in the wild collected in natural environments are extremely hard to annotate, which is mainly on account of the uneven quality of images from the Internet, as well as the complex backgrounds and occlusion interference. These uncertainties often lead to incorrect labeling and inconsistent labeling standards which will seriously affect the quality of datasets and bring difficulties to network training.

Before the rise of deep learning, traditional FER methods are mainly based on hand-crafted features (i.e., LBP \cite{shan2005lbp}, LDA \cite{deng2005new}, PCA \cite{mohammadi2014pca}, and SIFT \cite{ng2003SIFT}) following by the classifiers such as SVM \cite{platt1998SVM}. In recent years, with the development of parallel computing hardware, methods based on convolutional neural networks (CNNs) have gradually replaced the traditional methods and achieve state-of-the-art performance in FER tasks. However, the performance of these CNN-based approaches can be easily affected by the quality of the datasets and often lacks robustness, which is caused by the different backgrounds, image quality, head poses, and occlusion interference of the images. Especially, directly recognizing expressions on non-frontal faces caused by variant head poses is a big challenge \cite{zheng2014emotion}.

\begin{figure}[htb]
  \centering
  \centerline{\includegraphics[width=8.5cm]{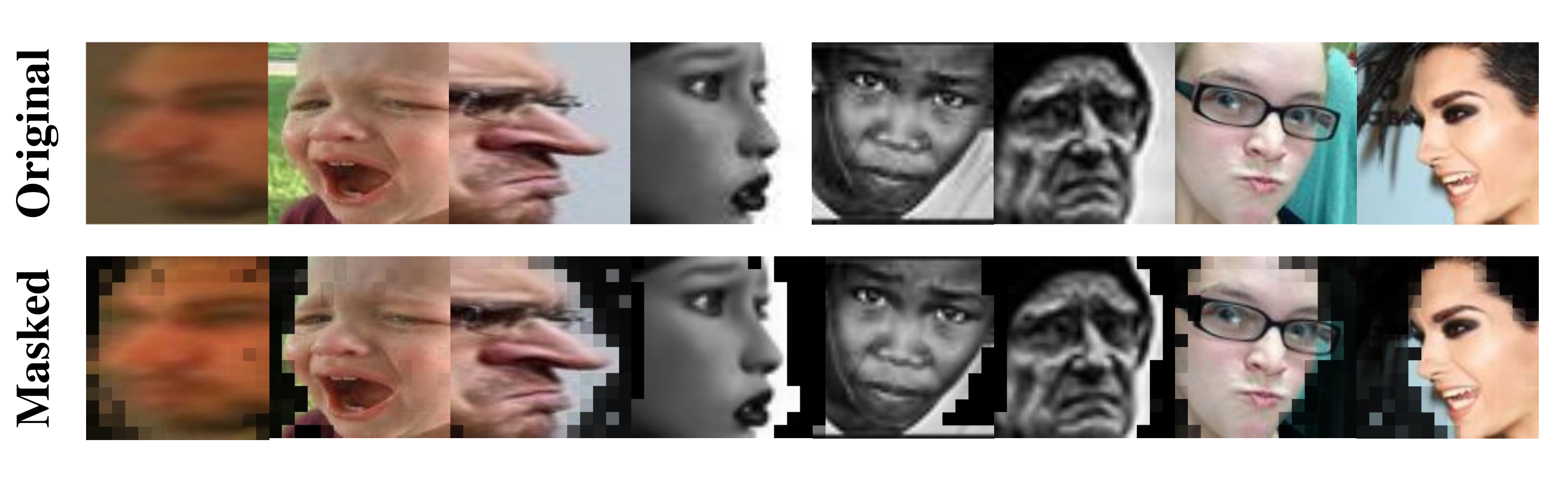}}
  \medskip

\caption{Samples from RAF-DB, FERPlus, and AffectNet. The top row shows the original images in the wild, while the bottom row shows the corresponding masked images. The black area represents the masked regions.}

\end{figure}
In the past few years, the transformer has become the model of choice in natural language processing (NLP) \cite{vaswani2017transformer}. Motivated by the success of transformer in NLP tasks, ViT achieves huge success in image classification tasks by pretraining ViT on a large-scale dataset and then fine-tuning on a smaller dataset \cite{dosovitskiy2020ViT}. DeiT further reduces the training costs of ViT by utilizing data enhancement and knowledge distillation \cite{touvron2020DeiT}. Back to the aforementioned FER problem in the wild, studies have shown that ViT has strong robustness against severe occlusion and disturbance \cite{naseer2021intriguing}. Therefore, transformer is particularly suitable to deal with FER in the wild. 

In this paper, we propose a convolutional-free model for FER termed as Mask Vision Transformer (MVT). To the best of our knowledge, it is the first pure transformer-based framework to solve the FER problems. By generating a mask for each facial image and improved relabeling strategy \cite{wang2020SCN}, our MVT effectively reduces the uncertainties and the impact of incorrect labels on the classification in FER, and achieves state-of-the-art or comparable performance on RAF-DB, FERPlus and AffectNet datasets. The proposed MVT consists of two crucial modules: dynamic relabeling and mask generation. Given a face image, a lightweight transformer-based network called mask generation network (MGN) is used to generate the mask for each image, which is employed to filter out the areas that are not relevant to expression recognition. As we can see in Fig. 2, the mask can effectively wipe off the useless background areas and occlusion interference which are harmful to FER. The training process of MGN is similar to generative adversarial network (GAN) \cite{goodfellow2014GAN}, which will be described in detail in Section 3. Then the masked image is split into several patches, linearly embed each of them, add position embeddings, and feed the resulting sequence of vectors to a ViT. After learning the relationship between patches with global self-attention mechanism, we utilize the class token to decide the expression. The second module is a relabeling module that relabels the suspected incorrect labels by comparing the maximum predicted probabilities to those of the given labels. Different from the fixed threshold in the relabeling module in SCN \cite{wang2020SCN}, we propose a dynamic relabeling module where the threshold varies with the probabilities of the given labels. Through the experiment, we find that the relabeling strategy with a fixed threshold can sometimes lead to erratic training because triggering the relabeling operation is usually caused by the incomplete fitting of the network instead of the incorrect label when the probabilities of the given labels are relatively high. Therefore, the relabeling with the fixed weights is easy to lead to the instability in training process. The proposed dynamic relabeling effectively makes the training process more stable and improves the performance on public datasets.

In summary, this paper has the following contributions:

\begin{itemize}

\item[$\bullet$] We propose a pure transformer-based model called MVT. To the best of our knowledge, MVT is the first FER framework built entirely based on transformer. Our
MVT achieves state-of-the-art results on RAF-DB, FERPlus, AffectNet-7 datasets and a comparable result on AffectNet-8 dataset.

\item[$\bullet$] We propose a mask generation network (MGN) inspired by GAN, which is totally different from most traditional GANs used to generate real-world images. The mask generated by our MGN can effectively filter out the backgrounds and interference of face images, retaining the expression information parts and boosting the classification accuracy. In addition, we propose a novel variance loss function to train our MGN.

\item[$\bullet$] We propose a new relabeling strategy, which makes the relabeling more accurate and the training process more stable, thus improving the performance on FER datasets in the wild.

\end{itemize}

\section{Related Work}
\subsection{Facial Expression Recognition in the Wild}
\begin{figure*}[htb]
  \centering
  \centerline{\includegraphics[width=15cm]{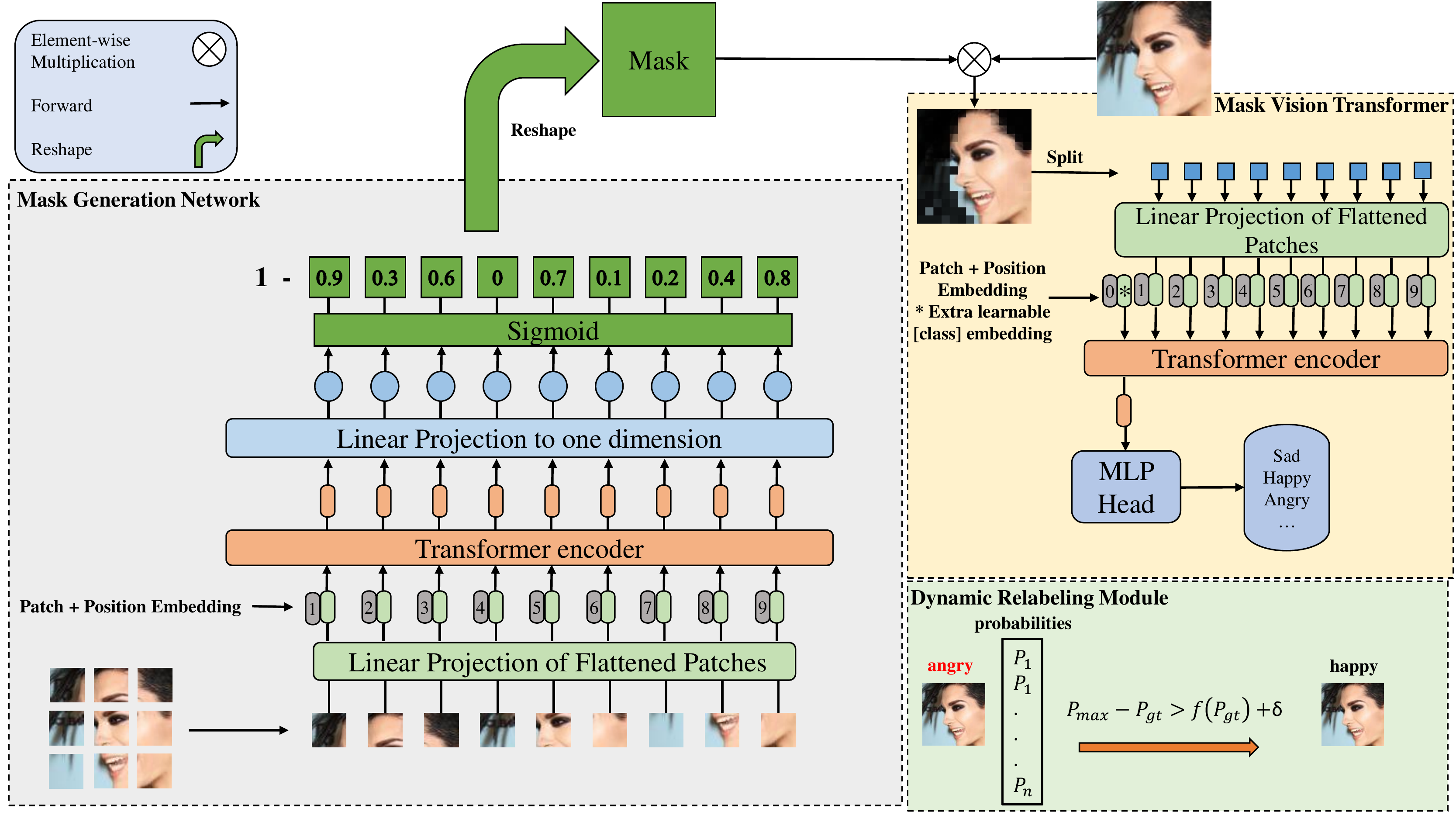}}
  \medskip

\caption{The pipeline of the pure transformer-based Mask Vision Transformer. It can be divided into three parts, mask generation network, mask vision transformer, dynamic relabeling module. We choose DeiT-S \cite{touvron2020DeiT} pre-trained on ImageNet as our classifier. The face images are fed to MGN to generate the corresponding masks which can measure the importance of the different areas and filter out complex backgrounds and occlusion. Then we utilize MVT to determine the expression of the masked images. During the training process, we design a dynamic relabeling module to rectify the suspected incorrect labels. }

\end{figure*}
In general, the existing FER methods are generally composed of two parts, face alignment and expression recognition. In the face alignment stage, MTCNN \cite{zhang2016MtCNN} and RetinaFace \cite{deng2020retinaface} are used to locate the face. The detected faces can be further aligned. In the expression recognition stage, hand-crafted features occupied the mainstream in FER at the beginning including SIFT \cite{ng2003SIFT}, HOG \cite{dalal2005HOG}, LBP \cite{shan2005lbp} and NMF \cite{buciu2004nmf,zhao2008pcanmf}. Deep learning based methods gradually replaced hand-crafted features in recent years. CNNs have a natural inductive bias for image processing. By making use of this advantage, many CNN-based methods have shown promising performance on lab-collected FER datasets \cite{li2017dlp-cnn,li2018gACNN,wang2020RAN,zhao2021efficientface}. However, the CNN-based models can be easily distributed by complex backgrounds, variant head poses, occlusion, etc. Wang \textit{et al}. \cite{wang2020SCN}, proposed a self-cure Network (SCN) to suppress the uncertainties for FER in the wild. Recent studies have shown that ViT has strong robustness against severe occlusion and disturbance \cite{naseer2021intriguing} which is the reason why we choose ViT as the backbone of our MVT.
\subsection{Vision Transformer}

Transformers were proposed by Vaswani \textit{et al.} \cite{vaswani2017transformer} for machine translation, and have become the state-of-the-art method in many NLP tasks. Inspired by the success of transformers, several researchers have tried to invest transformers in computer vision tasks, such as image classification \cite{dosovitskiy2020ViT,touvron2020DeiT}, object detection \cite{carion2020DETR}, segmentation \cite{wang2021pyramidtr}, etc. ViT is the first work to apply a vanilla transformer to image classification. By pre-training ViT on a large dataset (e.g. ImageNet \cite{krizhevsky2012imagenet}, ImageNet-21k \cite{deng2009imagenet}) and then fine-tuning on a smaller dataset (e.g. CIFAR-10/100 \cite{krizhevsky2009CIFAR10/100}), ViT sets new state-of-the-art on several classification datasets. DeiT further reduces the training costs of ViT by utilizing data enhancement and a distillation token \cite{touvron2020DeiT}. Both ViT and DeiT show superior performance compared with CNN-based methods. Inspired by ViT and DeiT, we first propose a pure transformer-based model for FER in the wild.
\subsection{Transformer based Generative Adversarial Network}

Deep generative models of images are neural networks trained to output synthetic imagery. At first, GAN can only generate low-resolution faces or numbers \cite{goodfellow2014GAN}. The latest generation models already can generate sample images that are hard for humans to distinguish from real-world images \cite{karras2020stylegan}. Inspired by ViT, scholars recently began to build generative networks based on transformers \cite{wodajo2021deepfake,jiang2021transgan}. Instead of generating real-world images, we propose a mask generation network based on transformer that generates masks for face images.

\section{Method}
To learn robust facial expression features with uncertainties, we propose a pure transformer-based framework Mask Vision Transformer (MVT). In order to fully demonstrate the advantages of ViT for visual modeling tasks, we retain the entire internal structure of ViT \cite{dosovitskiy2020ViT}. Since the training of ViT often requires large-scale data, we directly use the DeiT-S pretrained on ImageNet \cite{krizhevsky2012imagenet} in \cite{touvron2020DeiT}. Firstly, we provide an overview of the MVT, and then present its two main modules.

\subsection{Overview of Mask Vision Transformer}
Our MVT is built upon pure transformer and composed of two crucial modules: i) mask generation network, ii) dynamic relabeling, as depicted in Fig. 3.
\begin{figure*}[htb]
  \centering
  \centerline{\includegraphics[width=15cm]{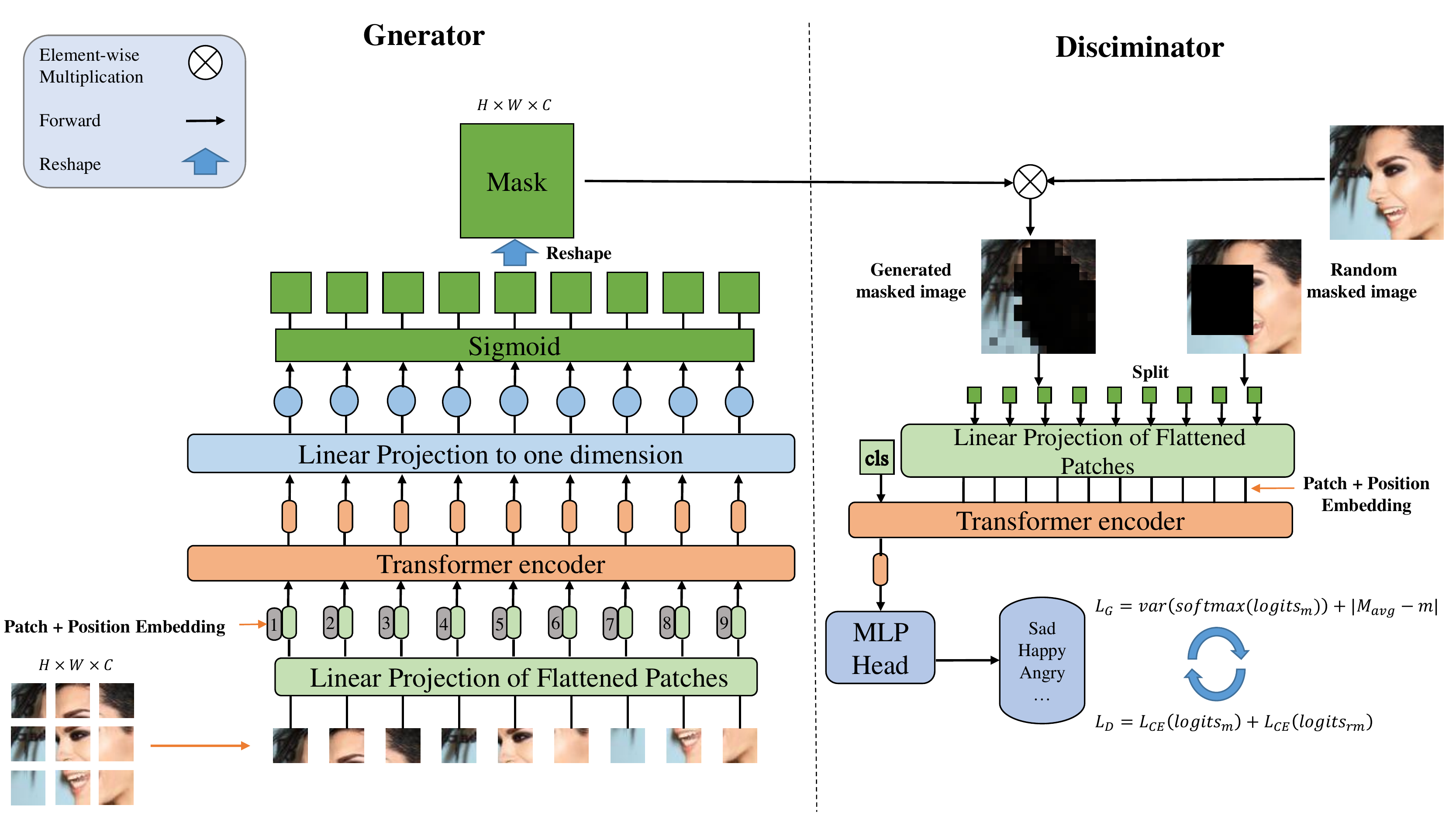}}
  \medskip

\caption{The pipeline of the pure transform-based generator and discriminator of MGN. Here $H = W = 224$ and $C=3$. }

\end{figure*}

For a given face image I$_{in}$ with the size of $H$$\times$$W$$\times$$C$, we reshape I$_{in}$ into a sequence of flattened 2D patches I$_{p}$ with the size of $N\times$$(P^{2}\times C)$, where $(H,W)$ is the resolution of the original image, $C$ is the number of channels, $(P,P)$ is the resolution of each image patch, and $N=H \times W/P^{2}$ is the number of patches which also serves as the effective input sequence length for our MGN. The MGN uses constant latent vector size $D$ through all of its layers, so we flatten the patches and map them to vector Z$_p$ with $D$ dimensions through a trainable linear projection. Then trainable position embeddings E$_{pos}$ with the size of $N\times D$ are added to the embedded patches Z$_p$ to retain positional information. The embedding vectors with the size of $N\times D$ are fed to MGN. In this paper, $D=384$ , $H=W=224$ and $P=16$. MGN outputs $N$ vectors with the size of $D$ after modeling the relationship between embedded patches. The outputs of MGN are then mapped to $N$ vectors with 1 dimension $M_{p_{i}}$ ranging from 0 to 1 with another trainable linear projection and sigmoid function. These values represent the importance of the corresponding patches to FER. Specifically, the higher the value is, the more unimportant this patch is. Since these values are negatively correlated with the importance of the patches, we need to subtract $M_{p}$ from 1 to get the patch masks which can filter out the backgrounds and occlusion of the face image. Then $M_{p_{i}}$ is copied to $P \times P \times C$ to match the size of input patches. These patch masks are restored to "images" in the spatial dimension based on the original positions to get the final mask M$_{f}$ with the size of $H$$\times$$W$$\times$$C$, which is the same as the input facial image I$_{in}$. We then do a dot product between M$_{f}$ and I$_{in}$ to get the masked image I$_{M}$, then I$_{M}$ is split into patches and fed to MVT to obtain the probabilities of each expression. It should be noted that only the parameters of MVT are optimized during the training, while the parameters of MGN are frozen. To further improve MVT, the dynamic relabeling module is added to rectify the samples with suspected incorrect label. This relabeling operation aims to hunt more clean samples and then enhance the final model. Different from the relabeling module from SCN \cite{wang2020SCN}, our relabeling threshold changes dynamically with the probabilities of given labels which can make the training process more stable and the module more robust.
\subsection{Mask Generation Network}

Inspired by the success of generative adversarial networks \cite{jiang2021transgan}, we propose a pure transformer-based mask generation network to generate a mask that can filter out the complex backgrounds and interference for each image. Instead of generating a real-world image, MGN generates the mask that acts as a filter which is different from most traditional GANs. To train our mask generation network, we need an extra ViT as a discriminator. As can be seen in Fig. 4, similar to the traditional GAN training mode, we train our MGN by alternately training generator and discriminator. When training the discriminator, we fix the parameters of MGN. Images with the mask generated by MGN and random masked images are fed to the discriminator for inference, and the loss function of the discriminator is formulated as:

\begin{equation}\label{1}
L_{D}=L_{CE}(logits_{gm})+L_{CE}(logits_{rm})
\end{equation}
\noindent where $L_{CE}$ is the cross-entropy loss commonly used in classification tasks; $logits_{rm}$ and $logits_{gm}$ are the output of MLP head of random masked image and generated masked image, respectively. By minimizing $L_{D}$, the discriminator is capable of recognizing expressions under masks. Specifically, the discriminator can learn the importance and relevance of different regions of the image to expression recognition by learning from random mask images which correspond to the real-world images in traditional GAN. Correspondingly, the generated masked images correspond to the generated image in the traditional GAN.

When training the generator of MGN, we fix the parameters of the discriminator. Only images masked by MGN are fed to the discriminator for inference. To train this special "GAN" we propose a novel generator loss function, formulated as:

\begin{equation}\label{2}
L_{G}=var(softmax(logits_{gm}))+\left | M_{avg}-m \right |,
\end{equation}
with
\begin{equation}\label{3}
var(x)=\sum_{i=0}^{K-1}\frac{(x_{i}-\bar{x})^{2}}{K-1},softmax(x)_{j}=\frac{e^{x_{j}}}{\sum_{i=0}^{K-1}e^{x_{i}}}
\end{equation}
and
\begin{equation}\label{4}
M_{avg}=\sum_{i=0}^{N-1}\frac{M_{p_{i}}}{N},
\end{equation}
\noindent where $\bar{x}$ is the mean of the elements of vector $x$, $x_{i}$ donates to the i-th element of the vector $x$ with $K$ dimensions, $M_{avg}$ is is the mean value of the generated patch masks $M_{p_{i}}$, and $m$ is the expected mask area size. We assume that when the generated masks can block the region related to the FER information, the discriminator will not be able to distinguish the expression of images. In other words, the discriminator will output each expression with equal probability. Therefore, by minimizing the variance of the expression probability, the MGN can learn the ability of measuring the importance of different areas of face images. The constant $m$ restricts the area of the generated mask, which can be flexibly adjusted according to different datasets. For the datasets with large backgrounds or interference areas, it can be set to a higher value. On the contrary, it can be reduced accordingly.

By training the generator and discriminator alternately, we can finally get a MGN that can generate a mask that covering the crucial regions for each face. In the inference stage, we only need to exploit the complement of the generated masks to effectively filter out the backgrounds and interference, so as to preserve the crucial areas needed for classification.

\subsection{Dynamic Relabeling}
Inspired by the relabeling module in SCN \cite{wang2020SCN}, we proposed the dynamic relabeling module to correct the mislabeled samples in FER datasets in the wild to improve the quality of the training sets. The original relabeling module chooses constant $\delta$ as the threshold to relabel the samples, that is to say, a sample is assigned to a new pseudo label if the maximum prediction probability is higher than the probability of the given label with a constant threshold $\delta$. The origin relabeling module is defined as
\begin{equation}\label{5}
l_{new}=\left\{\begin{array}{rcl} l_{max} & & if P_{max}-P_{gt}>\delta, \\l_{org}& & otherwise, \end{array}\right.
\end{equation}

\noindent where $l_{new}$ denotes the new label, $\delta$ is the threshold, $P_{max}$ is the maximum predicted probability, and $P_{gt}$ is the predicted probability of the given label. $l_{org}$ and $l_{max}$ are the originally given label and the index of the maximum prediction, respectively. We consider that it is unfair to use the same threshold as the criterion for relabeling when the probability of the given label is different. Specifically, when $P_{gt}$ is relatively high (e.g. $P_{gt}$ = 0.3), it may harm to relabel the sample with the index of the maximum prediction. By observing the relabeled samples in the experiment, we find that when $P_{gt}$ is relatively high, both the non-convergence of network and incorrect label can trigger a relabeling operation. When the samples are relabeled because the network does not converge, it is likely to change the correct label to the wrong one which can be extremely harmful to the network training. So we propose an improved relabeling module called dynamic relabeling. Formally, the module can be defined as,
\begin{equation}\label{6}
l_{new}=\left\{\begin{array}{rcl} l_{max} & &if P_{max}-P_{gt}>f(P_{gt})+\delta, \\l_{org}& & otherwise, \end{array}\right.
\end{equation}
\noindent where $f$ is a nonnegative monotone increasing function of $P_{gt}$ and $\delta$ is the lower limit of the threshold. This means that the relabeling threshold increases as $P_{gt}$ boosts, which effectively avoids the erroneous relabeling operation and makes the convergence more stable.
\section{Experiments}
To verify the effectiveness of the proposed method, we conduct the experiments on three popular in-the-wild facial expression datasets (i.e., RAF-DB \cite{li2017rafdb}, FERPlus \cite{barsoum2016ferplus} and AffectNet \cite{mollahosseini2017affectnet}). In this section, we first introduce the FER datasets used in our experiments and implementation details. We then explore the impact of each component of MVT on these datasets. Subsequently, we compare the proposed method with several state-of-the-art approaches.

\subsection{Datasets}

\textbf{RAF-DB} \cite{li2017rafdb} contains 29,672 facial images annotated with basic or compound expressions by 40 independent annotators. Consistent with most of the previous work, only images of seven prototypical expressions, such as neutral, happy, sad, surprised, fear, disgust, and anger, are used. Among them, 12271 images are used for training and 3068 images are for testing. The original size of the images in RAF-DB is 100 $\times$ 100. Both training images and test images have imbalanced distribution. The accuracy of the overall samples is executed for performance measurement.
\begin{table}[htb]
\label{1}
\caption{Evaluation of the mask ratio $m$ on FERPlus and RAF-DB. The best results are in bold.}
\begin{tabular}{  p{0.15\columnwidth}<{\centering} | p{0.35\columnwidth}<{\centering}  p{0.35\columnwidth}<{\centering}}
\toprule

 Datasets &Mask ratio (\%)&Accuracy (\%)\\
\midrule
       & 0 (Baseline)&86.7\\
RAF-DB & 10& 87.52\\
       & 15& \textbf{87.91} \\
       & 20&  87.45\\

\midrule
        &0 (Baseline)&87.92\\
FERPlus & 10& 88.45\\
        & 15&88.58 \\
        & 20& \textbf{88.88}\\

\bottomrule
\end{tabular}
\end{table}
\begin{table}[htbp]
\label{2}
\caption{ Comparisons between different dynamic relabel functions. The best results are in bold.}
\begin{tabular}{  p{0.15\columnwidth}<{\centering} | p{0.35\columnwidth}<{\centering}  p{0.35\columnwidth}<{\centering}}

\toprule

 Datasets &$f$&Accuracy (\%)\\
\midrule

       &Constant (Baseline)&86.86 \\
RAF-DB & Linear&87.09  \\
       & Quadratic&87.2  \\
       & Sigmoid  &\textbf{87.45}  \\
\midrule
        & Constant (Baseline)& 88.22\\
FERPlus & Linear& 88.32\\
        & Quadratic& \textbf{88.68}\\
        &Sigmoid& 88.52\\
\bottomrule
\end{tabular}
\end{table}

\textbf{FERPlus} \cite{barsoum2016ferplus} is extended from FER2013 employed in the \textit{ICML 2013 Challenges} in representation learning. FERPlus contains 28,709 training images, 3,589 validation images and 3,589 test images, which are all collected by the Google search engine. All the images are grayscale and have been resized to 48 $\times$ 48.  Each face image in FERPlus is annotated by 10 annotators. Apart from seven basic expressions as RAF-DB, contempt is included which leads to 8 expression classes.  We report the accuracy of all samples under the supervision of majority voting for performance evaluation.

\textbf{AffectNet} \cite{mollahosseini2017affectnet} is a large facial expression datasets with more than 1,000,000 face images gathered from the Internet. While only about 450, 000 images have been annotated manually with 11 expression categories. The seven expression categories denoted by AffectNet-7 involves neutrality, happiness, sadness, surprise, fear, disgust, and anger, while the eight expression categories denoted by AffectNet-8 with the additional contempt. It should be noted that AffectNet-7 and AffectNet-8 both have an imbalanced training set and a balanced test set. For AffectNet-7, there are 283,901 images as training data and 3,500 images for testing, and for AffectNet-8, there are 287,568 images as training data and 4,000 images as test data. We mainly present the mean class accuracy on the test set for performance measurement and fair comparisons with other methods.

\textbf{Occlusion and Pose Variant Datasets} is used to verify the performance of the FER model under real-world occlusion and variant head poses conditions. Wang \textit{et al.} \cite{wang2020RAN} built three subsets Occlusion-RAF-DB and Pose-RAF-DB from the test set of RAF-DB. The pose variations can be divided into two types, which are poses larger than 30 degrees and those larger than 45 degrees. We perform the accuracy of the overall samples for evaluation.

\subsection{Implementation Details}
In our experiments, images on all the datasets are resized to the size of 224 $\times$ 224. Then the backgrounds and occlusion of the images are masked by the corresponding masks generated by our MGN. It is worth noting that we do not perform any extra face alignment on all FER in the wild datasets. The random erasing, horizontal flipping, and color jittering are employed to avoid over-fitting. Since the transformer-based model benefits from pre-training on large-scale datasets, we use the Deit-S in \cite{touvron2020DeiT} pre-trained on ImageNet \cite{krizhevsky2012imagenet} as the backbone of both MGN and the classifier. In particular, we use the first six layers of DeiT-S with the pre-trained parameters as the backbone of MGN and the entire DeiT-S as the backbone of MVT. In the inference stage, We use AdamW \cite{loshchilov2018adamw} to optimize MVT with a batch size of 32 while keeping the parameters of MGN frozen. For RAF-DB and FERPlus, the learning rate is initialized to 0.00009, decreased at an exponential rate. For AffectNet-7 and AffectNet-8, the learning rate is initialized to 0.00001 with the same decay rate as RAF-DB and FERPlus. All the experiments are conducted on a single NVIDIA RTX 3070 card with Pytorch toolbox.

\subsection{Ablation Studies}
As shown in Fig. 3, our MVT mainly consists of mask generation network (MGN) and dynamic relabeling module. To show the effectiveness of our MVT, we conduct ablation studies to evaluate the influence of the key parameters and components on the final performance on RAF-DB and FERPlus datasets.

\textbf{Influence of the mask ratio.} We evaluate the recognition performance of MGN with different mask ratio $m$ in Eq. (2), as displayed in Table 1.

\begin{figure*}[htb]
\centering
\subfigure[]{
\begin{minipage}[t]{0.5\linewidth}
\centering
\includegraphics[width=3.6in]{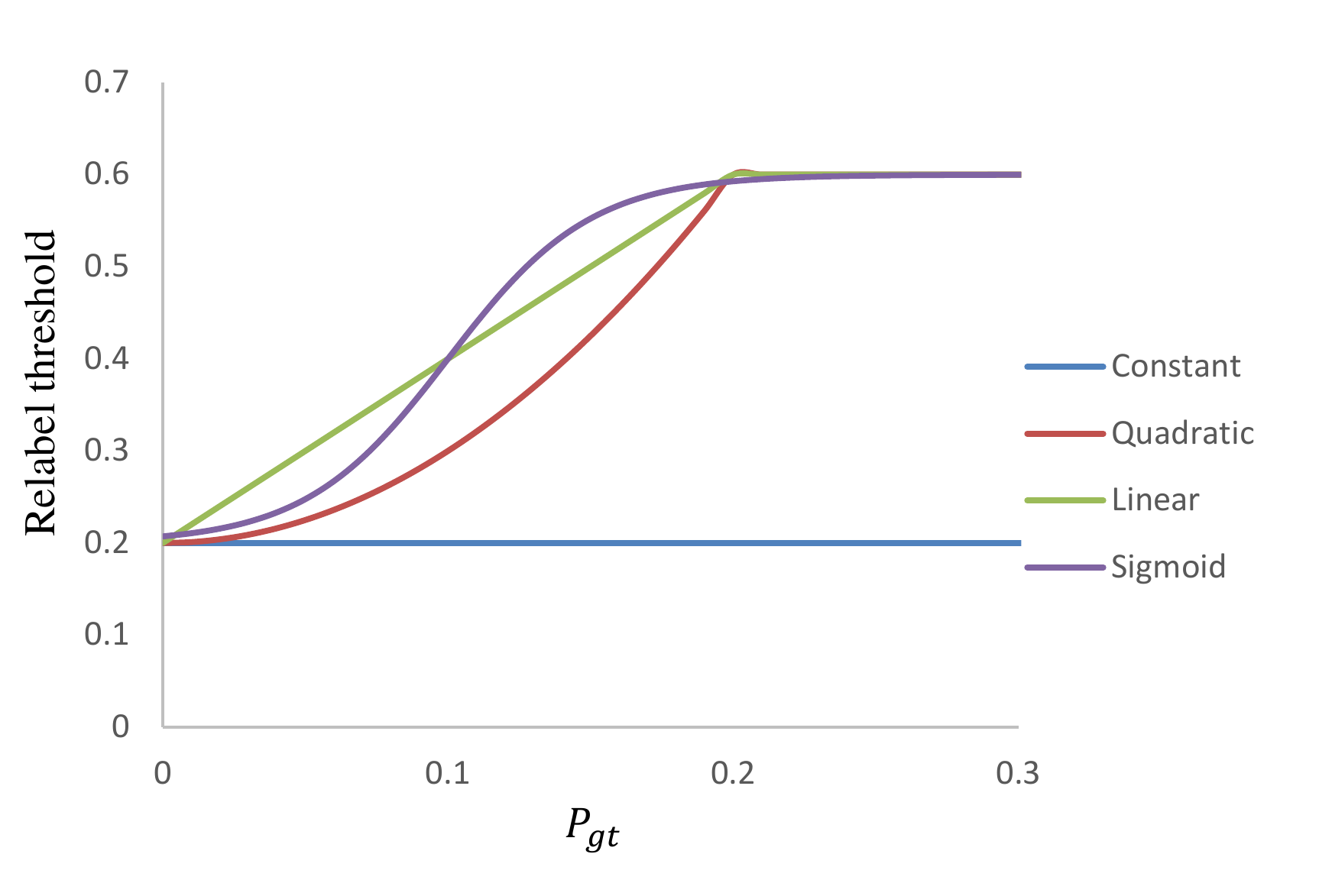}
\end{minipage}%
}%
\subfigure[]{
\begin{minipage}[t]{0.5\linewidth}
\centering
\includegraphics[width=3.6in]{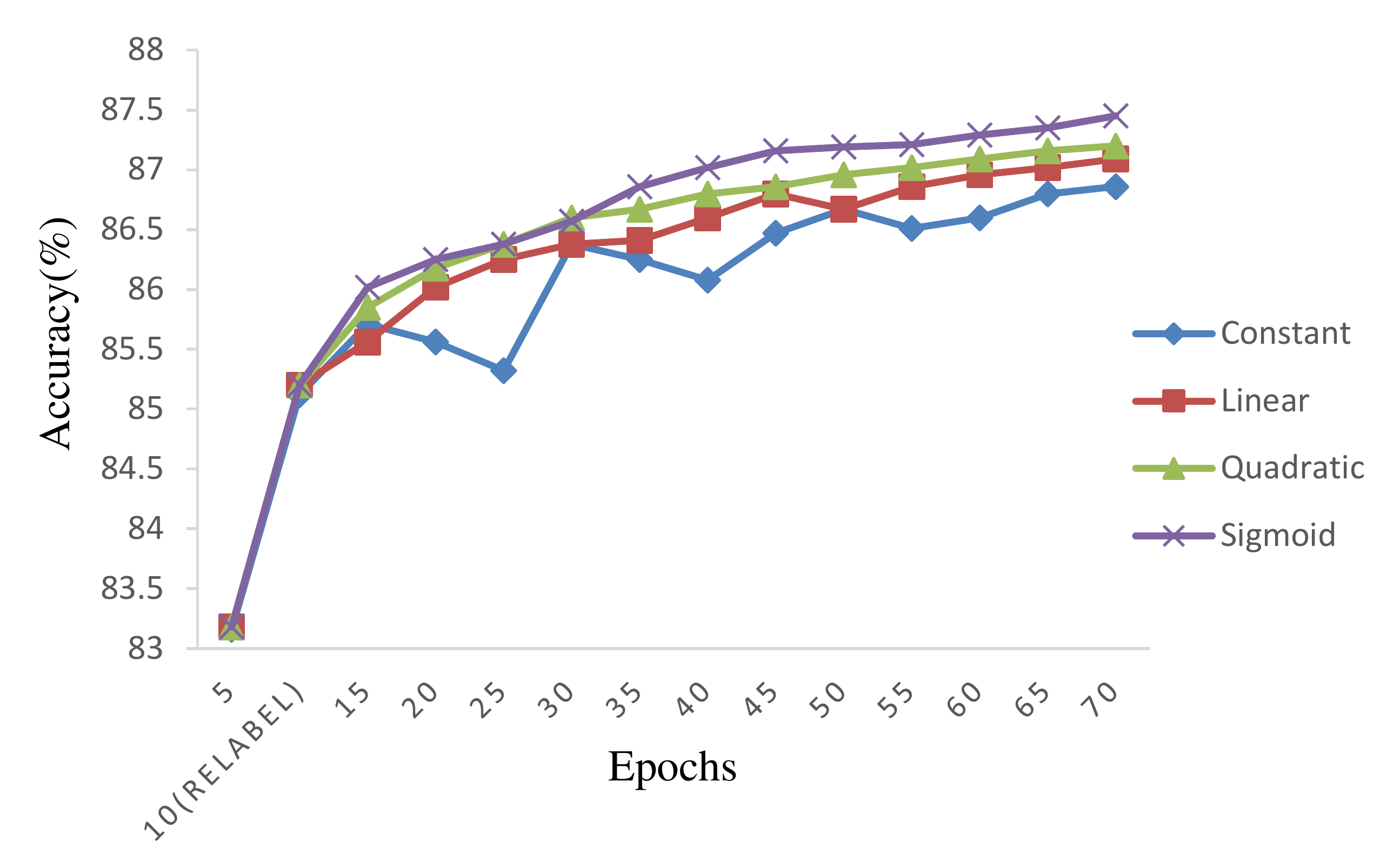}
\end{minipage}%
}%
\centering
\caption{Three dynamic relabeling threshold functions (Left: threshold function curves; Right: convergence curves of diferent $f$ on the RAF-DB test set).}
\end{figure*}
\begin{figure*}[htp]
\subfigure[Confusion Matrix of MVT on RAF-DB.]{
\begin{minipage}[t]{0.33\linewidth}
\centering
\includegraphics[width=2.3in]{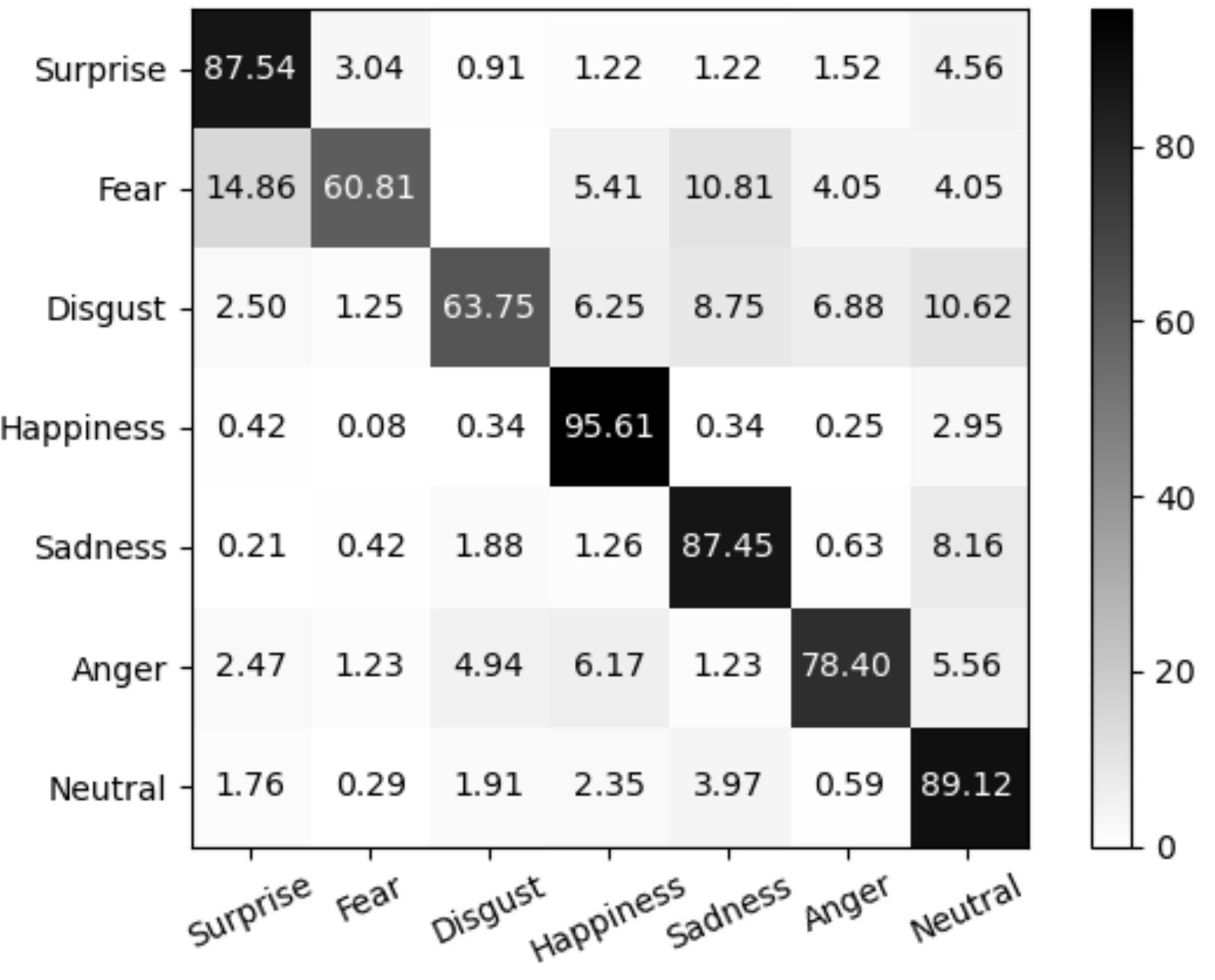}
\end{minipage}%
}%
\subfigure[Confusion Matrix of CVT on FERPlus.]{
\begin{minipage}[t]{0.33\linewidth}
\includegraphics[width=2.3in]{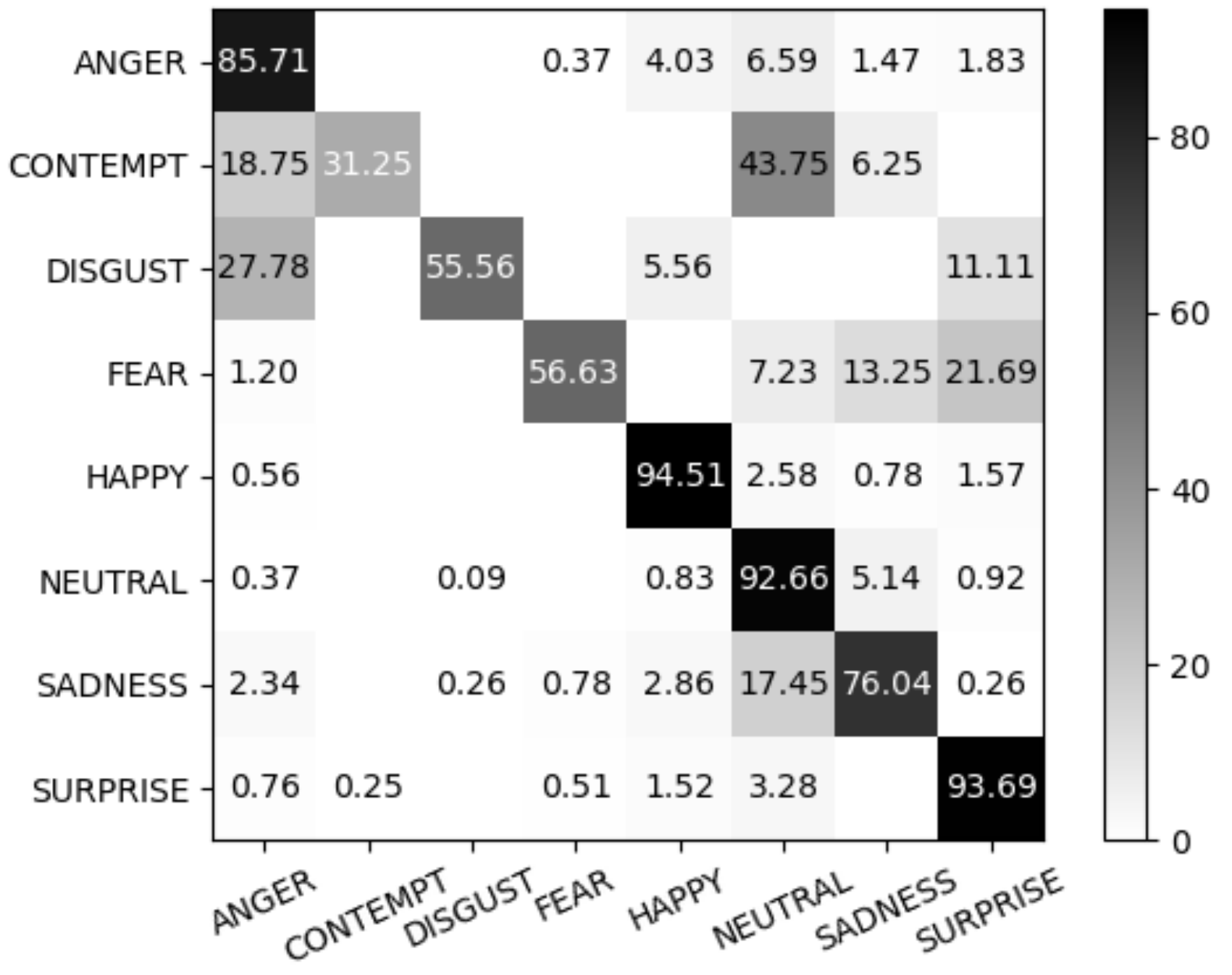}
\end{minipage}%
}%
\subfigure[Confusion Matrix of CVT on AffectNet-7.]{
\begin{minipage}[t]{0.33\linewidth}
\includegraphics[width=2.3in]{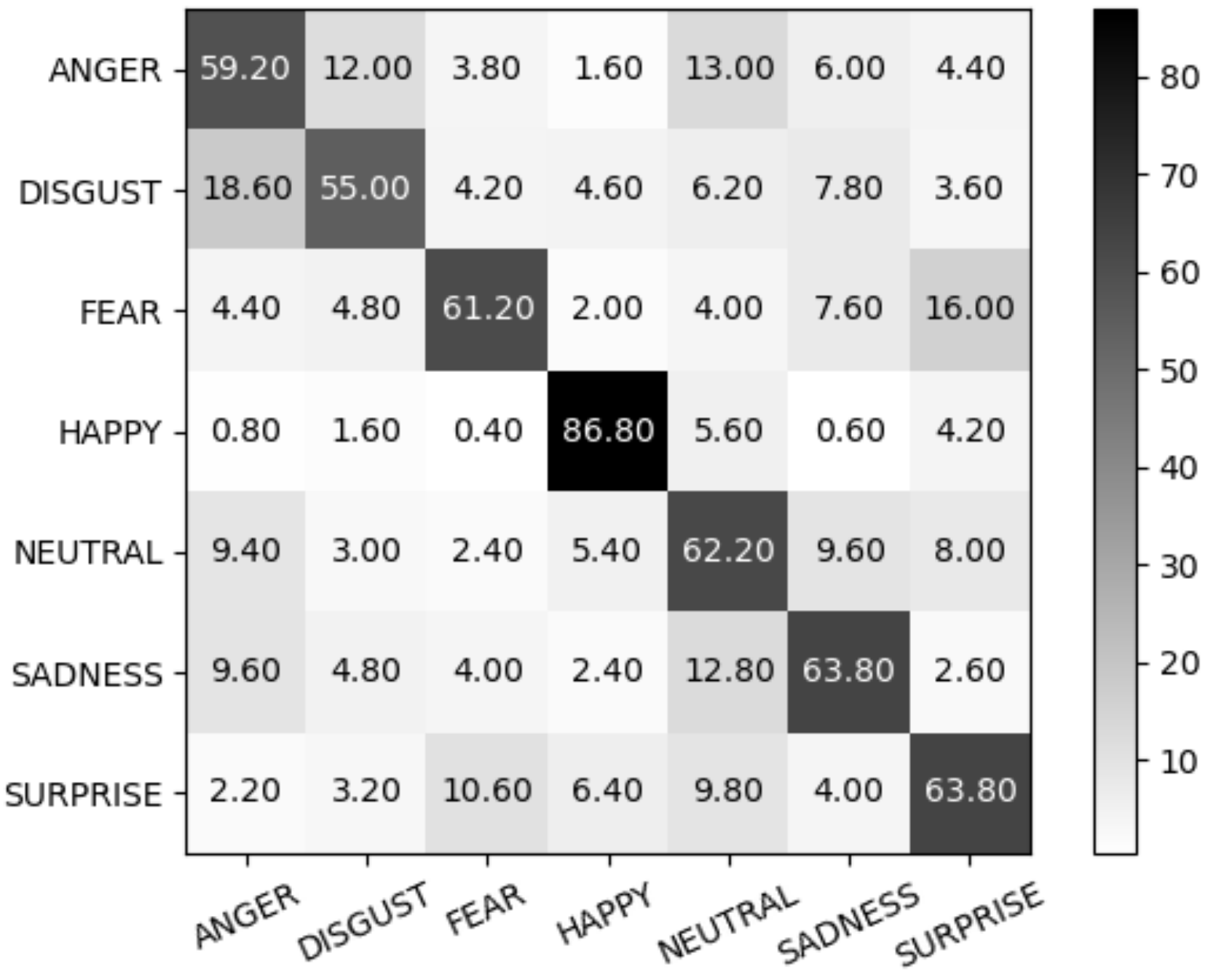}
\end{minipage}
}
\caption{The confusion matrices of our method on RAF-DB, FERPlus and AffectNet. (from left to right: RAF-DB, FERPlus, and AffectNet-7).}
\end{figure*}
We can observe that the optimal mask ratio is various for different datasets. Specifically, the optimal $m$ for RAF-DB is 15\% and is 20\% for FERplus. This is caused by the different proportions of backgrounds and occlusion areas in different datasets. As we aforementioned in Section 3, it is essential to train our MGN with various mask ratios according to different datasets. In the following, we set the values of $m$ to 15\% for RAF-DB, AffectNet-7, and AffectNet-8, while 20\% for FERPlus.

\textbf{Influence of Dynamic relabeling threshold function.} As we mentioned in Section 3, the dynamic threshold relabeling strategy can make the training process more stable, thus improving the performance. We first give three basic dynamic threshold functions of linear, quadratic, and sigmoid as pictured in Fig. 5(a). Specifically, we fix $\delta=0.2$ as Wang \textit{et al.} did in SCN \cite{wang2020SCN}. We also give the convergence curves of different $f$ on the RAF-DB test set with adding dynamic relabeling module and constant threshold relabeling module in 10-th epoch to prove the stability of the dynamic relabeling strategy as illustrated in Fig. 5(b).

From Fig. 5(b), it can be seen that the dynamic relabeling threshold is significantly better than the constant threshold in terms of both performance and convergence stability. This phenomenon is predictable because when relabeling threshold is the monotonically increasing function of $P_{gt}$, the problem of false relabeling caused by the non-convergence of the network can be effectively avoided. We then evaluate the recognition performance of dynamic relabeling with different types of $f$ in Eq. (6), as shown in Table 2.

We can find that the performance with different dynamic relabeling threshold functions is better than that of the constant threshold relabeling strategy in SCN on both RAF-DB and FERPlus datasets. In particular, it can be seen from that the optimal $f$ for RAF-DB is of the sigmoid type while for FERPlus is of the quadratic type from Table 2. We consider that this is because the images in FERPlus are grayscale images, which provide less information and lead to a smaller distance between classes of different expressions, a lower threshold is more suitable for FERPlus. In the following, we choose the sigmoid type $f$ for RAF-DB and the quadratic type $f$ for FERPlus.

\textbf{Effectiveness of Each Component in MVT.}
To verify the validity of MGN and dynamic relabeling module in our method, we conduct ablation studies on RAF-DB, FERPlus, and AffectNet-7 datasets. Specifically, the pre-trained DeiT-S \cite{touvron2020DeiT} is employed as the baseline in experiments, then MGN and dynamic relabeling module are added to the baseline, respectively. Experimental results are reported in Table 3.
\begin{table}[htbp]
\label{3}
\caption{ Evaluation of the MGN and dynamic relabeling modules. The best results are in bold.}
\begin{tabular}{  p{0.07\columnwidth}<{\centering}p{0.17\columnwidth}<{\centering}p{0.15\columnwidth}<{\centering}p{0.15\columnwidth}<{\centering}p{0.21\columnwidth}<{\centering}}

\toprule

 MGN&D-Relabel&RAF-DB&FERPlus&AffectNet-7\\
\midrule

\XSolidBrush & \XSolidBrush&86.67& 87.92&63.79 \\
\CheckmarkBold & \XSolidBrush&87.91 & 88.88& \textbf{64.57}\\
\XSolidBrush & \CheckmarkBold&87.45 & 88.68 & --\\
\midrule
\CheckmarkBold & \CheckmarkBold& \textbf{88.62} & \textbf{89.22}&--\\

\bottomrule
\end{tabular}
\end{table}
\begin{table}[htp]
\caption{ Comparisons with the state-of-the-art results. $^\circ$Extra face alignment is used. $^\dag$These results test on AffecNet-7. $^\ast$Oversampling is used
since the train set of AffectNet is imbalanced.}
\centering

\subtable[Comparisons with state-of-the-art methods on RAF-DB. The best results are in bold.]{

\begin{tabular}{p{0.45\columnwidth}<{\centering}  p{0.15\columnwidth}<{\centering} p{0.2\columnwidth}<{\centering} }
\toprule
Method & Year & Accuracy(\%)\\
\midrule
FSN \cite{zhao2018FSN}&2018&81.14\\
gACNN \cite{li2018gACNN}&2018&85.07\\
RAN \cite{wang2020RAN}&2020&86.90\\
SCN \cite{wang2020SCN}&2020&87.03\\
DSAN-VGG-RACE \cite{fan2020DSAN-VGG-RACE}&2020&85.37\\
SPWFA-SE \cite{li2020SPWFA-SE}&2020&86.31\\
CVT \cite{ma2021CVT}&2021&88.14\\
EfficientFace \cite{zhao2021efficientface}&2021&88.36\\
\midrule
Ours&2021&\textbf{88.62}\\
\bottomrule
\end{tabular}
\label{firsttable}
}

\qquad

\subtable[Comparisons with state-of-the-art methods on FERPlus. The best results are in bold.]{

\begin{tabular}{p{0.4\columnwidth}<{\centering}  p{0.2\columnwidth}<{\centering} p{0.2\columnwidth}<{\centering} }
\toprule
Method & Year & Accuracy(\%)\\
\midrule
ResNet+VGG \cite{huang2017ResNet+VGG}&2017&87.4\\
SHCNN \cite{miao2019SHCNN}&2019&86.54\\
LDR \cite{fan2020LDR}&2020&87.6\\
RAN \cite{wang2020RAN}&2020&87.85\\
RAN$^\circ$ \cite{wang2020RAN}&2020&88.55\\
SCN \cite{wang2020SCN}&2020&88.01\\
CVT \cite{ma2021CVT}&2021&88.81\\

\midrule
Ours&2021&\textbf{89.22}\\
\bottomrule
\end{tabular}
\label{firsttable}
}

\qquad

\subtable[Comparisons with state-of-the-art methods on AffectNet-7 and AffectNet-8. The best results are in bold.]{

\begin{tabular}{p{0.4\columnwidth}<{\centering}  p{0.2\columnwidth}<{\centering} p{0.2\columnwidth}<{\centering} }
\toprule
Method & Year & Accuracy(\%)\\
\midrule
IPA2LT$^\dag$ \cite{zeng2018IPA2LT}&2018&55.11\\
gACNN$^\dag$ \cite{li2018gACNN}&2018&58.78\\
SPWFA-SE$^\dag$ \cite{li2020SPWFA-SE}&2020&59.23\\
DDA-Loss$^{\dag\ast}$ \cite{farzaneh2020DDA-Loss}&2020&62.34\\
EfficientFace$^{\dag\ast}$ \cite{zhao2021efficientface}&2021&63.7\\
\midrule
RAN \cite{wang2020RAN}&2020&52.97\\
RAN$^\ast$ \cite{wang2020RAN}&2020&59.5\\
SCN$^\ast$ \cite{wang2020SCN}&2020&60.23\\
CVT$^\ast$ \cite{ma2021CVT}&2021&\textbf{61.85}\\
EfficientFace$^\ast$ \cite{zhao2021efficientface}&2021&59.89\\

\midrule
Ours$^\ast$&2021&61.40\\
Ours$^{\dag\ast}$&2021&\textbf{64.57}\\
\bottomrule
\end{tabular}
\label{firsttable}
}
\end{table}
\begin{table}[htp]
\label{5}
\caption{ Comparisons with other methods on Occlusion-RAF-DB and Pose-RAF-DB datasets}
\begin{tabular}{  p{0.35\columnwidth}<{\centering}  p{0.15\columnwidth}<{\centering}  p{0.15\columnwidth}<{\centering} p{0.15\columnwidth}<{\centering}}

\toprule

 Method &Occlusion&Pose(30) & Pose(45)\\
\midrule
Baseline \cite{wang2020RAN}& 80.19&84.04&83.15\\
RAN \cite{wang2020RAN}& 82.72& 86.74&85.2\\
CVT \cite{ma2021CVT}& 83.95& 87.97&88.35 \\
EfficientFace \cite{zhao2021efficientface}& 83.24&\textbf{88.13}&86.92\\
\midrule
Ours&\textbf{85.17}&87.99&\textbf{88.40}\\

\bottomrule
\end{tabular}
\end{table}

We can see that after filtering out the backgrounds and occlusion by MGN, the result exceeds the baselines by 1.24\%, 0.96\%, and 0.78\% on RAF-DB, FERPlus and AffectNet-7 datasets, respectively, which indicates the importance and validity of wiping of the occlusion and backgrounds for FER in the wild datasets. Moreover, by employing the dynamic relabeling module during the training process, our method also outperforms the baseline.

In particular, on account of the large amount of data in training set and the imbalanced data distribution in AffectNet, the training accuracy should not be too high (i.e. 65\% for AffectNet), otherwise it will lead to over-fitting. However, the relabeling module can only be added when the training accuracy is high enough (e.g. 90\%) so as to utilize features learned by the network to help correct the wrong labels. Therefore, we do not add the dynamic relabeling module in the training of AffectNet.

\subsection{Comparison with State-of-the-arts}
We compare our method with previous state-of-the-art methods on RAF-DB, FERPlus, AffectNet-7, and AffectNet-8. To our knowledge, we achieve new state-of-the-art results on RAF-DB, FERPlus, AffectNet-7, and a comparable result on AffectNet-8. Specifically, MVT outperforms recent state-of-the-art methods with 88.62\%, 89.22\%, and 64.57\% on RAF-DB, FERPlus, and AffectNet-7, respectively as presented in Table 4. We also give the confusion matrices on RAF-DB, FERPlus, and AffectNet-7 to show the superiority of our method as shown in Fig. 6.

\textbf{Results on RAF-DB:} Comparisons with other state-of-the-art methods are listed in Table 4(a). Among them, FSN \cite{zhao2018FSN} propose a feature selection mechanism to improve the performance of a CNN-based model. gACNN \cite{li2018gACNN} leverages a patch-based attention network and a global network. SPWFA-SE \cite{li2020SPWFA-SE} also does not report specific expression recognition accuracy, it provides the confusion matrix on RAF-DB. Therefore, we borrow the accuracy results from its confusion matrix for comparison. CVT \cite{ma2021CVT} is the first to introduce transformer into FER in the wild task even though they use CNNs to extract features. EfficientFace \cite{zhao2021efficientface} is a lightweight network with label distribution training. Our proposed method achieves 88.62\% on RAF-DB. From Table 4(a), it can be witnessed that the proposed method outperforms all of these state-of-the-art methods.

\textbf{Results on FERPlus:} Comparison with other state-of-the-art methods are shown in Table 4(b). Among them, SHCNN \cite{miao2019SHCNN}, ResNet+VGG \cite{huang2017ResNet+VGG}, LDR \cite{fan2020LDR}, RAN \cite{wang2020RAN}, and SCN \cite{wang2020SCN} are CNN-based method. Especially, RAN$^\ast$ means original RAN with extra face alignment. CVT \cite{ma2021CVT} utilize both CNNs and multi-layer transformer encoder to recognize expressions. Even without face alignment in our experiment settings, MVT still achieve the state-of-the-art results 89.22\% on FERPlus.

\textbf{Results on AffectNet:} We compare MVT with several state-of-the-art methods on AffectNet-7 and AffectNet-8, respectively in Table 4(c). For AffectNet-8, our MVT achieves a comparable result of 61.40\% with CVT \cite{ma2021CVT}. It can be seen that there is a large gap between the results of AffectNet-7 and AffectNet-8 which is mainly caused by adding expression of contempt based on AffectNet-7. As illustrated in \cite{zhao2021efficientface}, there exists lots of noise in the eighth expression categories, which can seriously deteriorate the accuracy. Our MVT outperforms state-of-the-art method \cite{zhao2021efficientface} by 0.87\%, which uses the label distribution for training.

\textbf{Results on Occlusion and Pose Variant Datasets:} To verify the robustness of the MGN to occlusion and variant head poses, we conduct several experiments on Occlusion-RAF-DB and Pose-RAF-DB which are collected by \cite{wang2020RAN} to examine methods under real-world scenario. Table 5 shows the accuracy on Occlusion-RAF-DB and Pose-RAF-DB with the same experimental setting as RAF-DB.

From Table 5, We can see that our MVT obtains the gains of 1.93\% and 1.22\% over EfficientFace and CVT on Occlusion-RAF-DB, which fully proves the effectiveness of our method for filtering out occlusion. In addition, we also achieves comparable results with state-of-the-art methods on Pose-RAF-DB.

\section{Conclusion}
In this paper, we firstly propose a pure transformer-based framework Mask Vision Transformer (MVT) for FER in the wild. Specifically, we present the mask generation network (MGN) which can filter out the complex backgrounds and occlusion to provide clean data for the inference. In the inference stage, based on the original constant threshold relabeling strategy, a novel dynamic relabeling module is proposed to hunt more clean labeled samples and enhance the final model. Extensive experiments on three public datasets in the wild demonstrate  that our MVT achieves new state-of-the-art results and is robust to real-world occlusion.
\section{REFERENCES}
\label{sec:ref}

\bibliographystyle{IEEEbib}
\bibliography{aaai}

\end{document}